%% file: 0_main.tex
\pdfoutput=1

\documentclass[hf]{ceurart}
\usepackage{enumitem}
\usepackage{subcaption}
\usepackage{hyperref}
\usepackage{listings}
\begin{document}

\copyrightyear{2024}
\copyrightclause{Copyright for this paper by its authors.
  Use permitted under Creative Commons License Attribution 4.0
  International (CC BY 4.0).}

\conference{IARML@IJCAI'2024: Workshop on the Interactions between Analogical Reasoning and Machine Learning, at IJCAI'2024,
  August, 2024, Jeju, South Korea}

\title{Enhancing Analogical Reasoning in the Abstraction and Reasoning Corpus via Model-Based RL}

\author[1]{Jihwan Lee}[
email=jihwan.lee@gm.gist.ac.kr
]
\fnmark[1]
\author[1]{Woochang Sim}[
email=woochang@gm.gist.ac.kr
]
\fnmark[1]
\author[1]{Sejin Kim}[
email=sejinkim@gist.ac.kr
]
\cormark[1]
\author[1]{Sundong Kim}[
email=sundong@gist.ac.kr
]
\cormark[1]
\address[1]{Gwangju Institute of Science and Technology}

\fntext[1]{Equal contribution.}
\cortext[1]{Corresponding authors.}

\input{1_abstract}


\maketitle

\input{1_intro}
\input{2_background}
\input{3_experiment}
\input{3.1_single_task}
\input{3.2_similar_task}

\input{4_discussion}
\input{5_relatedworks}
\input{6_conclusion}

\begin{acknowledgments}
This work was supported by the IITP (No. 2019-0-01842, RS-2023-00216011), AICA (HPC-AI) and the GIST (AI-based Research Scientist Project, HPC-AI) funded by the Ministry of Science and ICT, Korea. QCT also supported computing servers for this work.
\end{acknowledgments}

\bibliography{0_main}

\end{document}

%% file: 1_abstract.tex
\begin{abstract}
This paper demonstrates that model-based reinforcement learning (model-based RL) is a suitable approach for the task of analogical reasoning. We hypothesize that model-based RL can solve analogical reasoning tasks more efficiently through the creation of internal models. To test this, we compared DreamerV3, a model-based RL method, with Proximal Policy Optimization, a model-free RL method, on the Abstraction and Reasoning Corpus (ARC) tasks. Our results indicate that model-based RL not only outperforms model-free RL in learning and generalizing from single tasks but also shows significant advantages in reasoning across similar tasks.
\end{abstract}

%% file: 1_intro.tex
\section{Introduction}

\begin{figure}[h!]
    \centering 
    \includegraphics[width=0.71\linewidth]{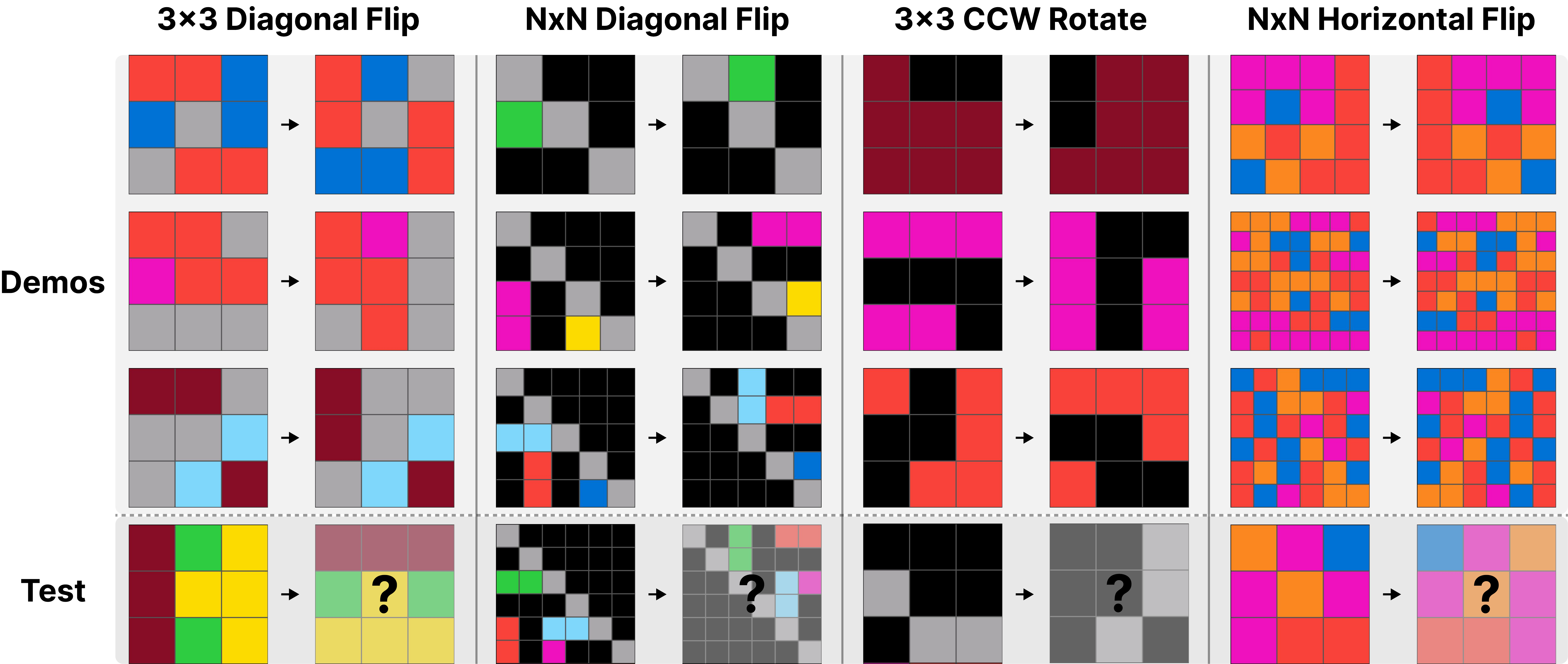} 
    \caption{Each ARC task includes several demos and a test input. The objective is to identify the grid corresponding to the test input by applying a common transformation rule found across all demos. The term ``CCW'' in the third task means counterclockwise.}
\label{fig:arc_example}
\end{figure}

The Abstraction and Reasoning Corpus (ARC)\footnote{ARC dataset URL: \href{https://github.com/fchollet/ARC-AGI}{https://github.com/fchollet/ARC-AGI}} is known as a benchmark for evaluating abstraction and reasoning abilities~\cite{malkinski2023review}.
Examples of ARC tasks can be seen in Fig.~\ref{fig:arc_example}. 
To solve an ARC task, one needs the ability to abstract a common rule among given demos and apply this rule to a new test input grid to infer the corresponding grid~\cite{chollet2019ARC}.
Additionally, to solve untrained tasks, one must have the ability to discover various knowledge during the learning process and selectively apply this knowledge in the task-solving process~\cite{chollet2019ARC}.

This study interprets the two abilities required by ARC not as separate skills but as one concept of analogical reasoning.
Analogical reasoning refers to the process of understanding new situations or solving problems based on existing knowledge that can be applied to different contexts or problems~\cite{hofstadter1979analogical}. 
First, the process of solving an ARC task can be seen as solving a new problem (test input grid) based on the transformations from input grids to output grids in each demo (existing knowledge), which exactly fits the definition described above.
Second, for a model trained on various ARC tasks to solve a new ARC task, it must understand the demos of the new ARC task (new situations) based on the solution processes (existing knowledge) applicable to the training tasks (different problems) and solve the task, which also aligns with the concept of analogical reasoning.
Third, since the evaluation tasks in ARC are typically more complex and challenging than the training tasks~\cite{bober2024ARClevel}, the ability to infer new problem-solving processes based on learned knowledge through analogical reasoning can enhance the performance on ARC.

In this study, noting the critical importance of analogical reasoning capabilities in solving ARC, we conducted reinforcement learning (RL) for training ARC tasks. 
Learning ARC with RL can offer the following potential benefits from an analogical reasoning perspective. First, RL can enhance its learning strategies through reward-based learning. 
In the process of learning ARC tasks, RL agents learn to maximize rewards across various tasks, thereby developing analogical reasoning capabilities. 
Second, RL can learn policies that effectively operate in new, yet similar environments based on experience. 
During the learning of various ARC tasks, RL agents can learn to apply rules learned in one scenario to another, a key element of analogical reasoning. 
Third, RL allows agents to produce their own data, enabling effective learning even without high-quality given data. 
In tasks like ARC, where data may be limited, RL agents can quickly infer patterns from a few demos and learn to apply these to the new input.

This study compared model-based RL and model-free RL to evaluate the efficiency of acquiring analogical reasoning abilities. 
The key difference between these two methodologies lies in the presence or absence of an internal environmental model.
Model-based RL builds an internal model of the environment based on the agent's experiences and learns policies through predictable scenarios, which has the potential to significantly improve decision-making in complex situations.
On the other hand, model-free RL learns to optimize rewards through direct interaction with the environment, allowing for the development of effective action policies without constructing an environmental model.

This study hypothesized that the internal model of model-based RL could aid in analogical reasoning.
To test this hypothesis, we observed the learning processes of ARC tasks using DreamerV3~\cite{hafner2023mastering}, a representative model-based RL algorithm, and Proximal Policy Optimization (PPO)~\cite{schulman2017proximal}, a general model-free RL algorithm, demonstrating the potential for RL agents to acquire analogical reasoning.
Additionally, we evaluated the performance and learning efficiency of both algorithms by fine-tuning them on tasks similar to the pre-trained tasks and comparing the analogical reasoning capabilities they acquired.

%% file: 2_background.tex
\section{Model-Based RL and Model-Free RL}

RL has two main categories: model-based RL and model-free RL. Before conducting experiments to compare these categories from the perspective of abstraction reasoning, this paper briefly reviews the characteristics of each category and compares the algorithms that will be used in this experiment.

\subsection{Model-Based RL}

Model-based RL builds an internal model of the environment to predict various possibilities and establish more efficient policies based on these predictions.
In problems requiring abstract thinking and pattern recognition like ARC, the model-based RL approach provides the ability to generalize to new situations that the agent has not experienced through its internal model. 
Representative algorithms utilizing the model-based RL approach include Dyna-Q~\cite{sutton1990dynaq}, I2A~\cite{racaniere2017i2a}, MBMF~\cite{nagabandi2018mbmf}, and World Model~\cite{ha2018world}. 
Also, AlphaZero~\cite{silver2017alphazero}, which successfully surpassed human performance in board games such as Go, Chess, and Shogi based on Monte Carlo Tree Search (MCTS), is the most well-known.
Recently, DreamerV3~\cite{hafner2023dreamerv3}, based on the World Model, has been the subject of extensive research.

A key feature of DreamerV3 is its ability to extract important features from input data and convert these into a latent representation, which is then used to learn a predictive model of the environment's dynamics. 
This model functions by taking the current state and actions as inputs and predicting the next state's representation and rewards. 
This approach allows for the simulation of future scenarios and the evaluation of various scenarios. 
In summary, DreamerV3 operates by repeating the following three processes: 1) collect data based on a random policy or the current policy, 2) update the agent's model based on the collected data, and 3) generate and evaluate virtual future scenarios using the updated model to improve the policy. This process excels in multi-step tasks like mining diamonds in Minecraft. We chose DreamerV3 for its superior handling of ARC, compared to other model-based approaches.

\subsection{Model-Free RL}

Model-free RL operates directly through interactions with the external environment, basing its action policies on reward information. 
This methodology utilizes only the agent's experiences without prior modeling of the environment, making it relatively simple to implement and fast to execute.
These characteristics could constrain the flexible reasoning and rapid adaptation required in ARC. 
Representative algorithms of model-free RL include DQN~\cite{mnih2013dqn}, PPO~\cite{schulman2017ppo}, and SAC~\cite{haarnoja2018sac}.

Among these, the Proximal Policy Optimization (PPO) algorithm is one of the most commonly used algorithms in RL, based on the policy gradient method. 
PPO is designed to achieve stable and efficient learning even in complex environments. 
Its process can be summarized similarly to DreamerV3 in the following two steps: 1) perform actions in the environment based on the current policy to collect data, and 2) incrementally improve the policy using the policy gradient technique based on the collected data.

%% file: 3_experiment.tex
\section{Experiments}

We conducted an experiment comparing the ARC task learning of DreamerV3 and PPO, which are representative algorithms of model-based RL and model-free RL, respectively, to demonstrate the differences in analogical reasoning capabilities between these methodologies. 
Through this experiment, we aimed to answer the following research questions (RQs):

\begin{enumerate}[label=RQ\arabic*., leftmargin=30pt, itemsep=0pt, parsep=0pt]
\item Can Model-Based RL Learn a Single Task? ($\mathcal{A}\rightarrow \mathcal{A}$)
\item Can Model-Based RL Reason about Tasks Similar to Pre-Trained Task? ($\mathcal{A} \rightarrow \mathcal{A'}$)
\item Can Model-Based RL Reason about Sub-Tasks of Pre-Trained Task? ($\mathcal{AB} \rightarrow \mathcal{A}$)
\item Can Model-Based RL Learn Multiple Tasks Simultaneously? ($\mathcal{A, B} \rightarrow \mathcal{A}$)
\item Can Model-Based RL Reason about Merged-Tasks of Pre-Trained Tasks? ($\mathcal{A, B} \rightarrow \mathcal{AB}$)
\end{enumerate}

We scope our paper to provide answers on RQ1 and RQ2.
By addressing RQ1 and RQ2 through experiments, we aim to demonstrate the abilities of model-based RL in efficiently learning and applying knowledge across similar tasks.
The following descriptions are the common setting of every experiment.

\paragraph{Action}
The vast action space of ARC is one of the biggest obstacles to the RL~\cite{chollet2019ARC}.
According to the ARCLE framework~\cite{lee2024arcle}, action is defined as a combination of operation and selection; where operation means the type of action to be performed, and selection determines grids to which these actions are applied.
We force strict restrictions on these operations and selections to weaken the challenge posed by the vast action space. 
In experiments, RL agents can select only five operations: Rotate90, Rotate270, FlipH, FlipV, and Submit~\cite{lee2024arcle}, and selections are always fixed as the entire grid. 
These limitations influenced the selection of tasks.

\paragraph{Task}
We found tasks from among the 400 training tasks that could be solved with the restricted operations and selections of actions, and 7 tasks satisfied this condition. 
Among these tasks, we selected 4 tasks for our experiment as shown in Fig.~\ref{fig:arc_example}.
First, we selected tasks that need diagonal flipping on a $3 \times 3$ grid or $N \times N$ grids.
These tasks need the right combination of actions (rotate and flip) before submission.
Thus, we thought these tasks were good for measuring analogical reasoning ability.
Next, we chose tasks that need to rotate a $3 \times 3$ grid and flip horizontally on $N \times N$ grids.
These tasks need only one action except submit, so analogical reasoning ability are not necessary for agents. 

\paragraph{Reward}
We designed the reward based on a sparse reward such that an agent gets a reward of 1000 if it does the submit action and the state matches the correct grid. 
Additionally, even if the submit action is not executed, reaching the correct grid grants an extra reward of 1. 
Although sparse rewards can present challenges in learning tasks with a wide search space, we overcame these difficulties through simple tasks and restricted action spaces. 
Furthermore, by giving a small reward for just finding the correct grid, the RL agent can learn the importance of achieving the goal state, and by offering a large reward for the submit action, we prevented the agent from repeatedly reaching the correct grid without submitting in a single episode.

\paragraph{Metric}
In all experiments, we measured the data efficiency of learning through the accuracy relative to the number of environment steps used during training. 
We decided to use environment steps to ensure consistency with the experimental settings of DreamerV3~\cite{hafner2023dreamerv3}, as we wanted to monitor data efficiency throughout the learning process. 
Additionally, we employed the \textit{pass@3} method used in ARC to measure accuracy~\cite{chollet2019ARC}. 
The term \textit{pass@3} grants three submission opportunities per episode, and an episode is considered successful if the correct answer is found within three attempts. 
If three incorrect submissions are made or the predetermined episode length (50) is exceeded, the attempt is judged unsuccessful. 

\paragraph{Training}
All tasks used in the experiments were trained using the RL environment for ARC, ARCLE~\cite{lee2024arcle}. 
Each task was originally composed of a few demos and one test input; however, there was a concern that this could lead to overfitting during the learning process.
Therefore, we augmented the training with 1,000 demos and used an additional 100 test inputs for evaluating the model. 
Furthermore, when training a single task, the agent was trained over 100,000 environment steps, whereas for fine-tuning a new task on a pre-trained model, the agent was trained for 50,000 environment steps.

%% file: 3.1_single_task.tex
\subsection{Can Model-Based RL Learn a Single Task? ($\mathcal{A}\rightarrow \mathcal{A}$)}

\begin{figure}[ht]
    \centering
    \begin{subfigure}[b]{0.49\columnwidth}
        \includegraphics[width=\linewidth]{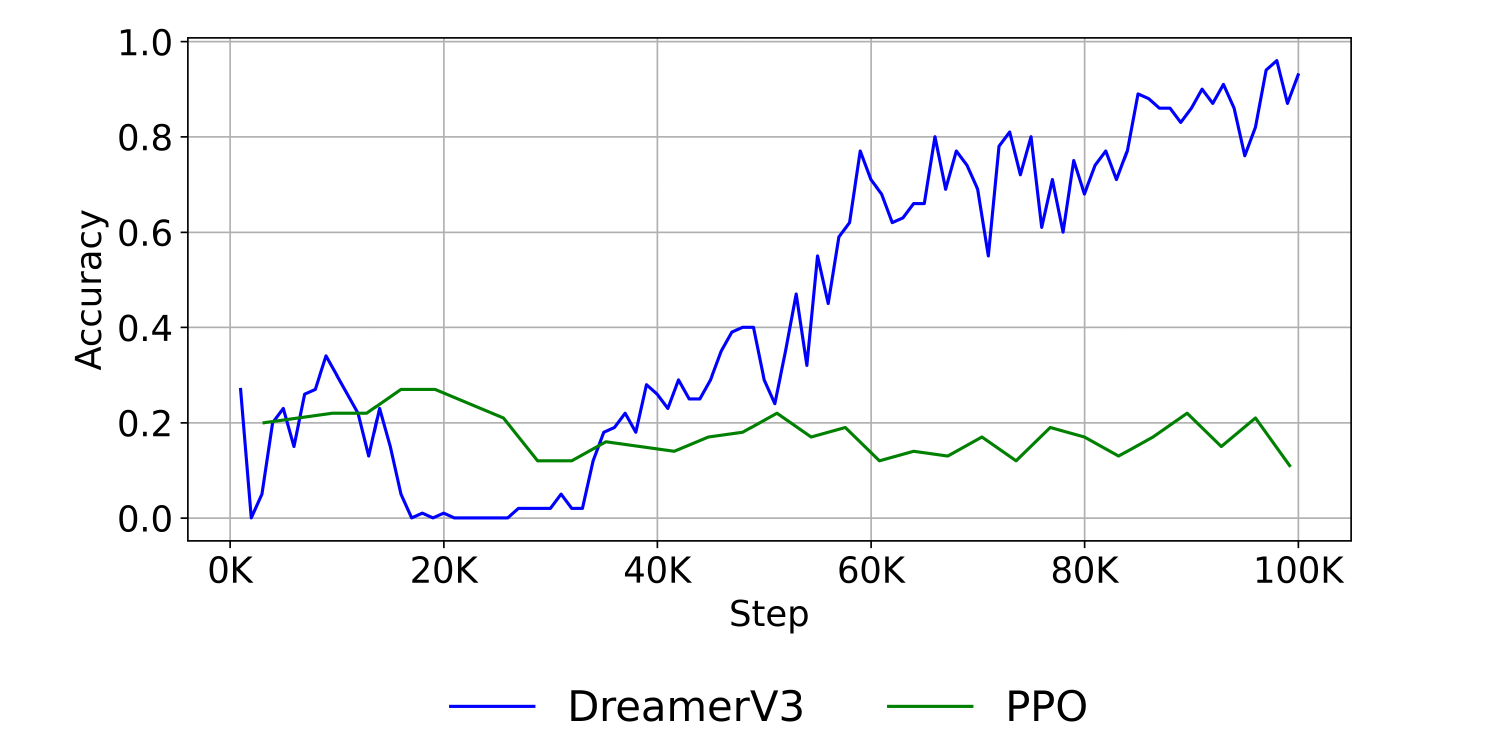}
        \caption{$3 \times 3$ Diagonal Flip}
        \label{fig:exp1_FlipD_3}
    \end{subfigure}
    \hfill 
    \begin{subfigure}[b]{0.49\columnwidth}
        \includegraphics[width=\linewidth]{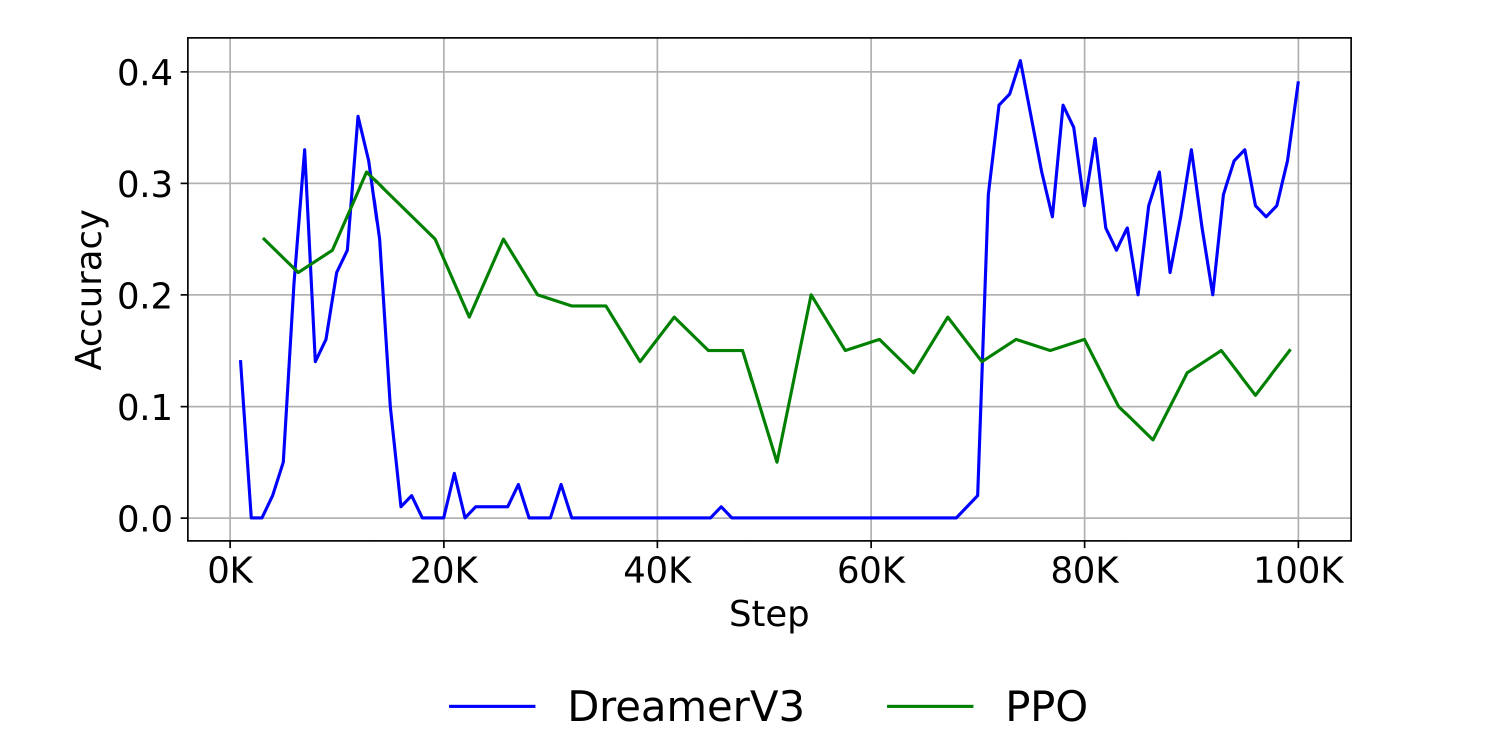}
        \caption{$N \times N$ Diagonal Flip}
        \label{fig:exp1_FlipD_N}
    \end{subfigure}

    \begin{subfigure}[b]{0.49\columnwidth}
        \includegraphics[width=\linewidth]{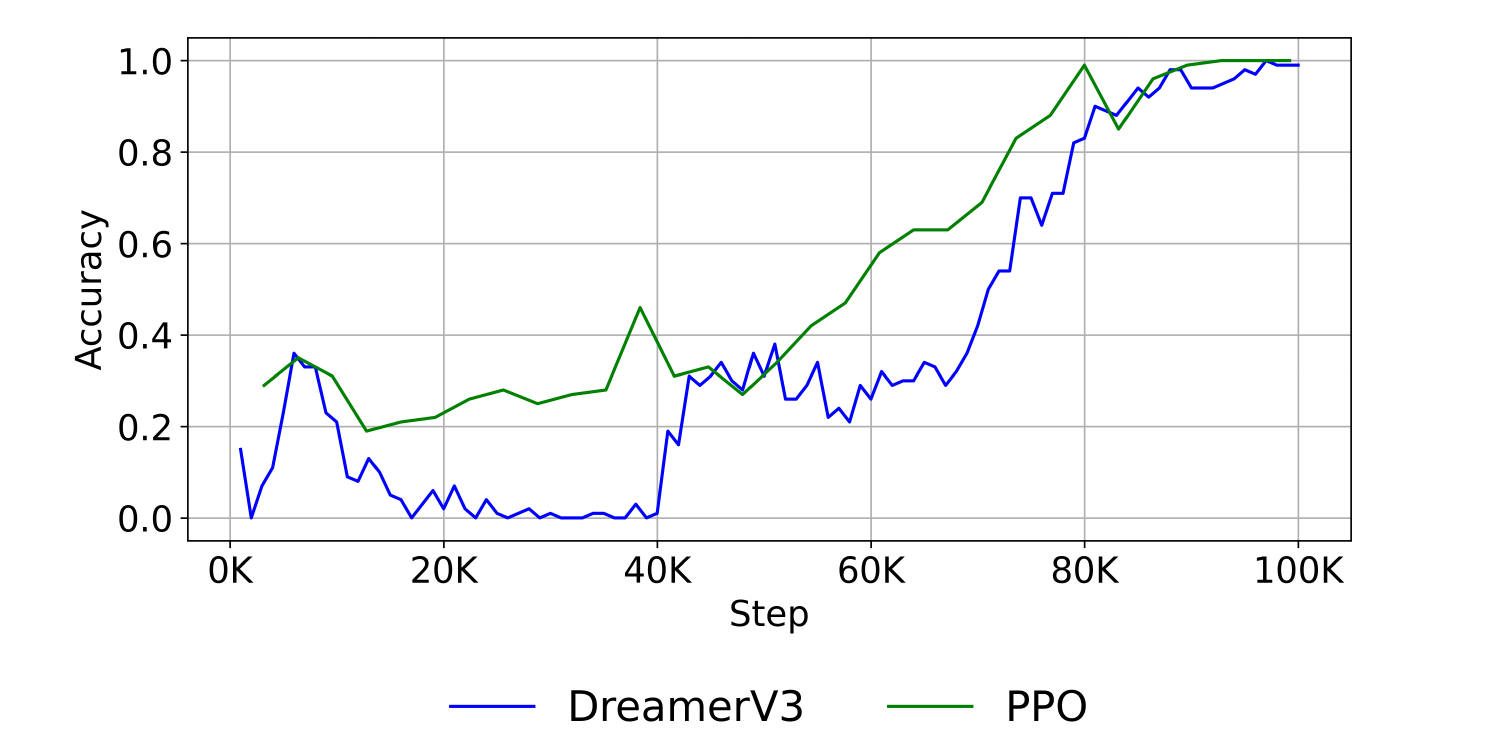}
        \caption{$3 \times 3$ CCW Rotate}
        \label{fig:exp1_Rotate270}
    \end{subfigure}
    \hfill 
    \begin{subfigure}[b]{0.49\columnwidth}
        \includegraphics[width=\linewidth]{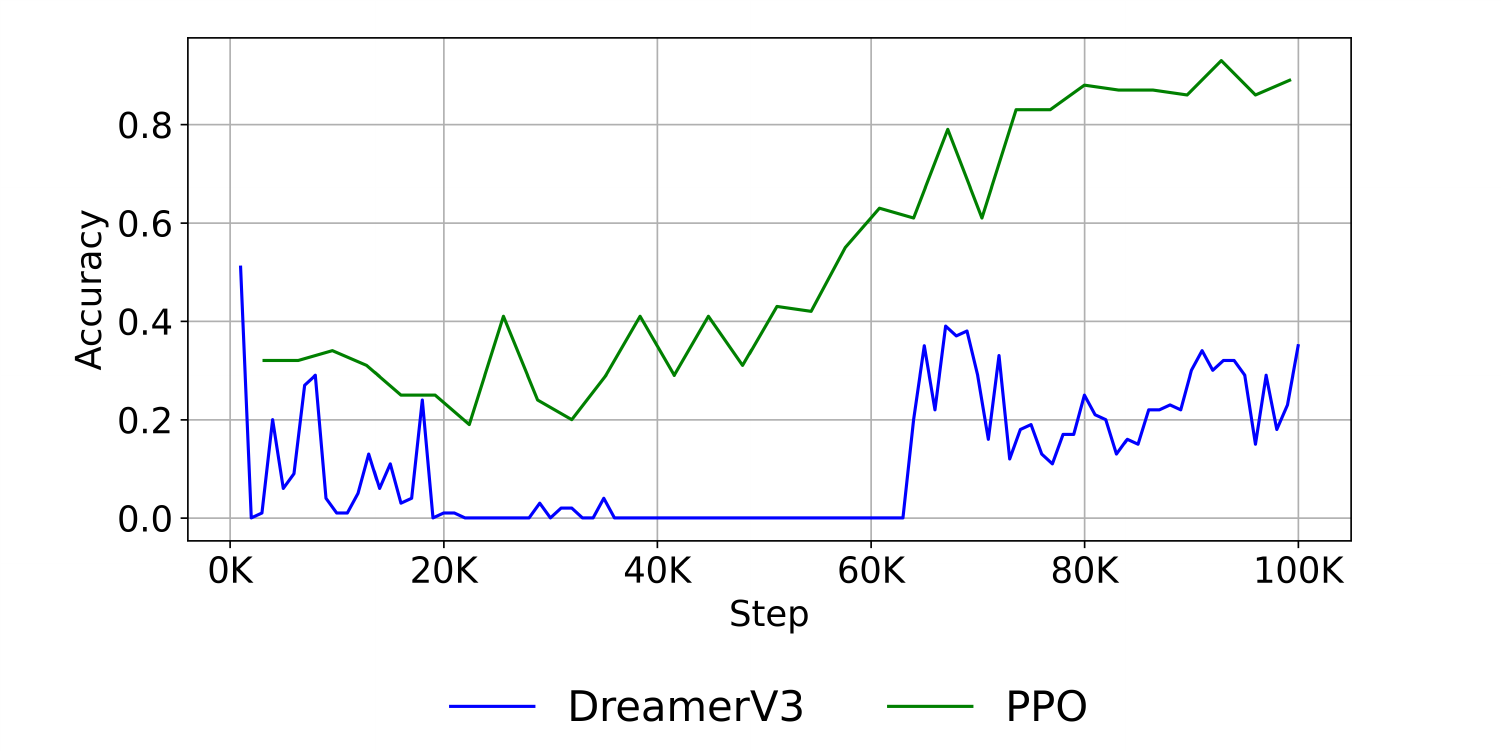}
        \caption{$N \times N$ Horizontal Flip}
        \label{fig:exp1_FlipH}
    \end{subfigure}

    \caption{Performance of agents on four single ARC tasks with different RL algorithms. The above two results show that the model-based RL agent learned better on analogical reasoning tasks. The below two results show that the model-free RL agent could be better on simple tasks. Additionally, an interesting common result was shown in the learning curve of model-based RL: there always occurs an interval in the middle of learning where accuracy drops to 0. We argue that this interval is where model-based RL learns concepts for analogical reasoning.}
    \label{fig:exp1}
\end{figure}

The first experiment focused on a single ARC task requiring analogical reasoning, where the objective was to learn the common rules from given demo pairs and apply these rules effectively to a test input. This experiment involved training on four ARC tasks, each illustrated in Fig.~\ref{fig:arc_example}, using the DreamerV3 and PPO algorithms.
Tasks requiring multiple actions to solve, such as Diagonal Flipping, demand analogical reasoning, whereas tasks that can be solved with a single action, such as CCW Rotation and Horizontal Flipping, are less complex.
DreamerV3, a model-based RL algorithm, exhibited superior performance in Diagonal Flip tasks, which involve more complex reasoning. 
In contrast, PPO, recognized for its stable learning capabilities, performed comparably or even better in simpler tasks.

It is noteworthy that DreamerV3 achieved 100\% performance on uniformly sized $3 \times 3$ tasks but only managed about 40\% on tasks with varying sizes.
Conversely, PPO's performance remained consistent regardless of grid size and was influenced solely by task difficulty. 
This suggests a performance drop in DreamerV3 during the encoding processes.

A common interval of zero performance was observed in DreamerV3's training across various tasks, followed by rapid performance improvement. 
This pattern suggests that DreamerV3 is not merely learning simple problem patterns but is developing a deeper understanding of concepts essential for analogical reasoning. 
Initially, DreamerV3 demonstrated about 40\% effectiveness due to learning straightforward patterns. 
Later, the agent recognized that this approach was insufficient for fully solving tasks, prompting it to explore and learn from various trials. 
Consequently, DreamerV3 displayed significant performance improvements in later learning stages, outperforming earlier results. 
These observations imply that DreamerV3's latent representations capture conceptual content that enhances learning efficiency and illustrates the potential for analogical reasoning within model-based RL frameworks.

%% file: 3.2_similar_task.tex
\subsection{Can Model-Based RL Reason Tasks Similar to Pre-Trained Task? ($\mathcal{A} \rightarrow \mathcal{A'}$)}

\begin{figure}[ht]
    \centering
    \begin{subfigure}[b]{0.49\columnwidth}
        \includegraphics[width=\linewidth]{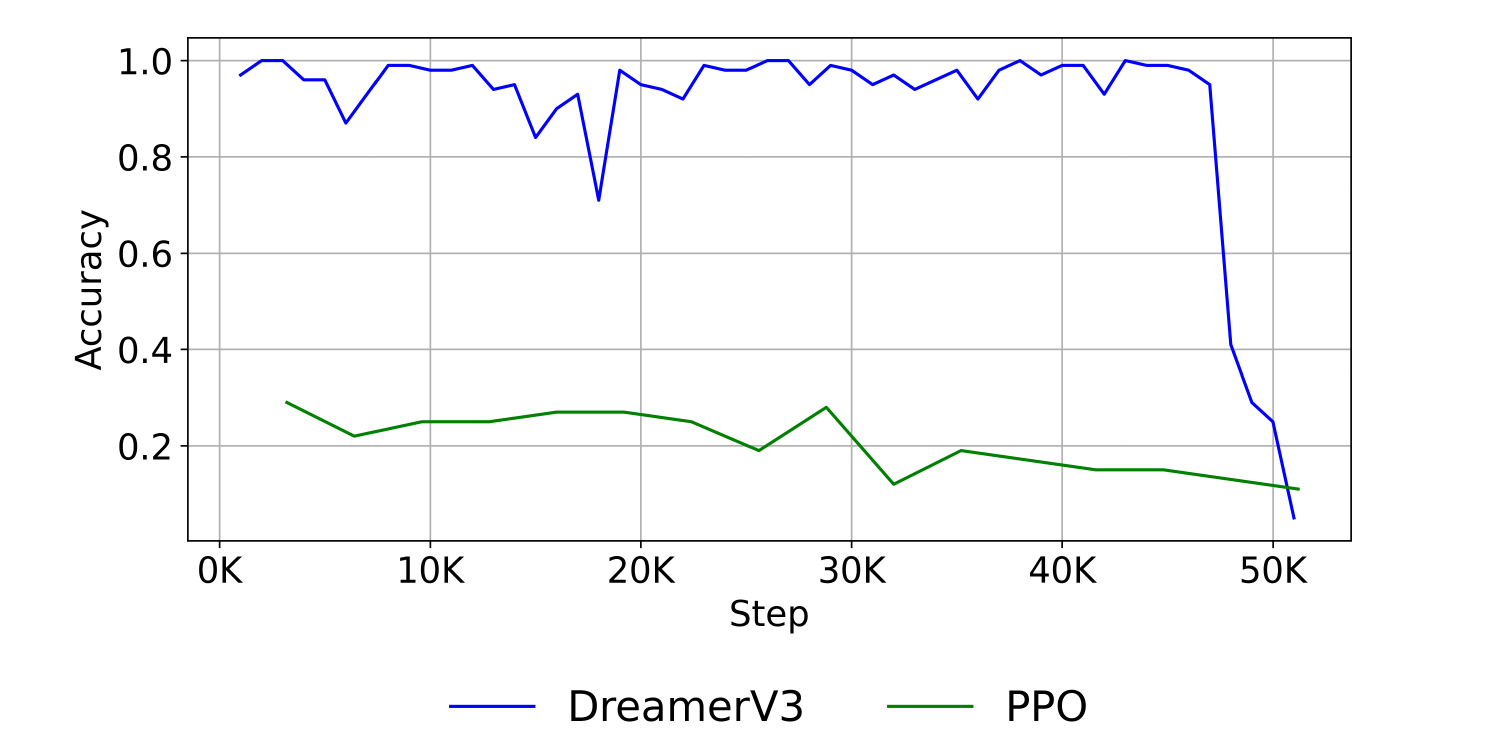}
        \caption{$3 \times 3$ to $N \times N$ Diagonal Flip}
        \label{fig:exp2_3toN}
    \end{subfigure}
    \hfill 
    \begin{subfigure}[b]{0.49\columnwidth}
        \includegraphics[width=\linewidth]{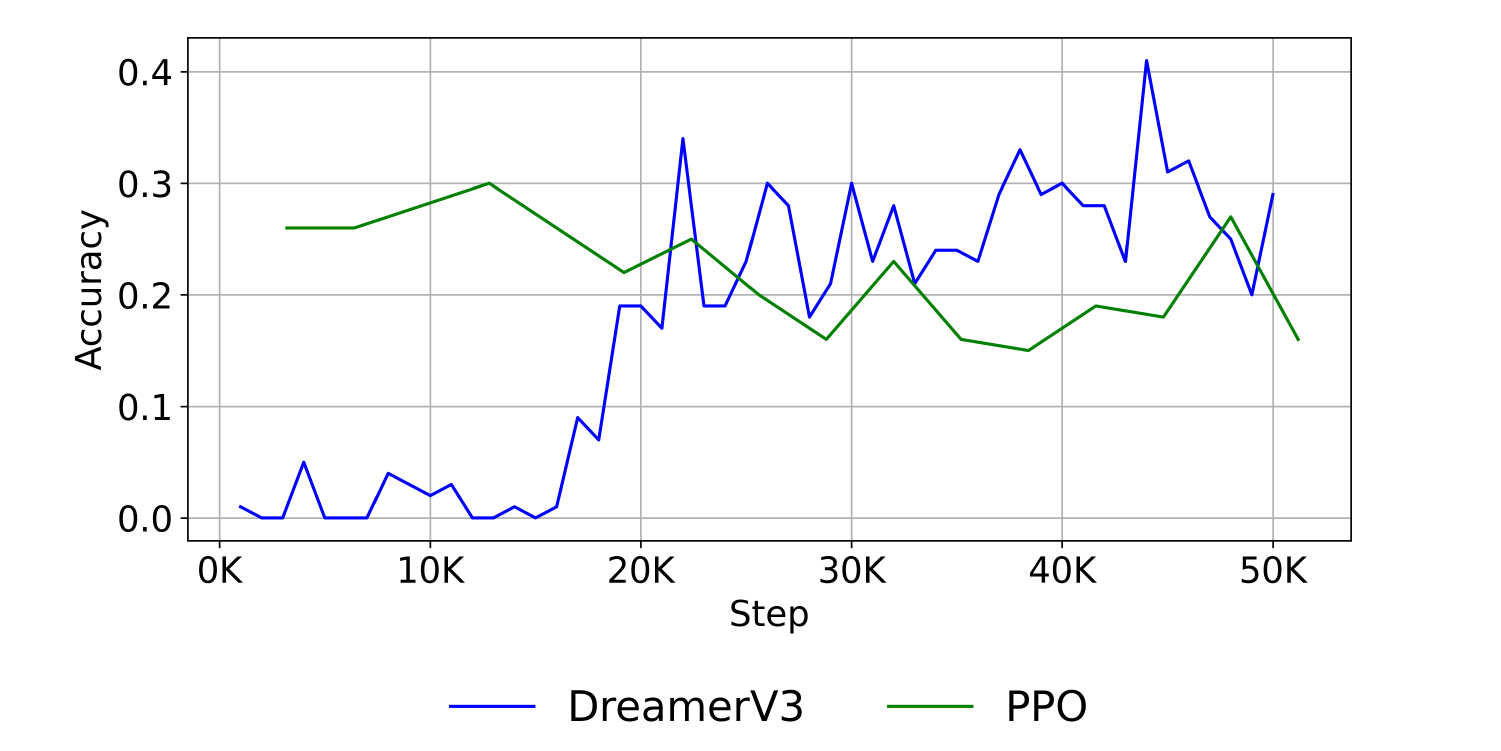}
        \caption{$N \times N$ to $3 \times 3$ Diagonal Flip}
        \label{fig:exp2_Nto3}
    \end{subfigure}
    \caption{Comparing the performance of agents between DreamerV3 and PPO on two single ARC tasks with a pre-trained model about a similar task. Understandably, PPO did not gain any benefit from fine-tuning due to the low performance of the pre-trained model. In contrast, DreamerV3 showed very high performance when adapting from a pre-trained model that had performed well in the $3 \times 3$ Diagonal Flip task. However, when utilizing a pre-trained model with poor performance, DreamerV3 also displayed lower initial learning efficiency than when no pre-trained model was used. Lastly, at the end of the experiment where fine-tuning was successful, a sudden drop in performance occurred, which is presumed to be the same phenomenon as the interval in previous experiments.}
    \label{fig:exp2}
\end{figure}

The second experiment evaluated the learning efficiency of an agent when facing a task similar to one it was previously trained on.
For instance, in Fig.~\ref{fig:exp2_3toN}, we observed the performance of diagonal flip tasks across various grid sizes for agents pre-trained on a $3 \times 3$ diagonal flip task. 
Conversely, in Fig.~\ref{fig:exp2_Nto3}, we reversed the roles of the pre-trained and adaptation tasks to evaluate both agents' performances.

In both scenarios, PPO consistently displayed a performance level of around 20\%. 
This outcome was anticipated, considering that the agent's performance on pre-trained tasks had similarly been around 20\%, as evidenced in previous experiments (Fig.~\ref{fig:exp1_FlipD_3} and Fig.~\ref{fig:exp1_FlipD_N}).
The insufficient performance of the pre-trained model likely inhibited any positive influence on the adaptation process, resulting in no notable performance gains compared to the non-pre-trained scenario.

An interesting finding from the DreamerV3 results was the direct impact of the pre-trained model's performance on adaptation. 
For example, while the model without pre-training achieved about 40\% performance on the same task, the pre-trained model approached 100\%, mirroring the $3 \times 3$ Diagonal Flip task's success previously encountered.
This pattern was further supported by the performance contrasts observed between Fig.~\ref{fig:exp1_FlipD_3} and Fig.~\ref{fig:exp2_Nto3}.
Here, a task that once achieved 100\% performance dropped to around 40\% after pre-training, reflecting the outcomes seen with the $N \times N$ Diagonal Flip tasks.
These outcomes illustrate that DreamerV3 tailored its problem-solving strategies based on insights drawn from pre-trained tasks, suggesting that better results could have been obtained if adaptation in Fig.~\ref{fig:exp2_Nto3} had also been initiated from a higher-performing pre-trained model.

Towards the latter part of training in Fig.~\ref{fig:exp2_3toN}, a sudden decline in performance was noted. 
While the exact cause of this drop cannot be definitively established, we hypothesize that it might be due to the same type of interval observed in previous results (Fig.~\ref{fig:exp1}).
Although the experiment's step limit prevented precise verification, it seems probable that the DreamerV3 agent was undergoing an interval similar to earlier tests, attempting to learn new concepts. 
Further experimentation beyond 50,000 steps and additional theoretical analysis are required to explore this phenomenon more deeply, which could further substantiate DreamerV3’s capability for analogical reasoning.

%% file: 4_discussion.tex
\section{Discussion}
\paragraph{Restriction of Action Space} Our study examined the analogical reasoning abilities of reinforcement learning (RL) algorithms and their ability to apply learned concepts. 
In the ARCLE, an agent's actions consist of operation and selection, requiring the policy to adeptly handle both components. 
However, the challenge of learning the selection process significantly increases the complexity of the tasks.
To mitigate this, we simplified the action space by providing selections through an oracle in this experiment. 
This adjustment allowed the agents to focus more on mastering operations without the added difficulty of selection, streamlining the learning process.
Future research should explore the possibility of incorporating the selection process more autonomously within the reasoning tasks. 
This would test the agent's ability to handle the complete process of decision-making, potentially leading to more analogical reasoning ability for agents.

\paragraph{Restriction of Tasks} This paper focused on learning simple tasks involving just one or two operations, such as diagonal flipping tasks, a Rotating task, and a horizontal flipping task. 
Future research should explore additional challenges such as the Rotate and the Flip task, which could provide valuable insights into the model's ability to generalize and reason about sub-components of learned tasks. 
It is also crucial to assess the efficiency of learning multiple related actions (multi-tasking) simultaneously. 
To overcome this challenge, previously learned knowledge must be retained while new knowledge is acquired. 
In other words, it is essential to be robust against catastrophic forgetting. 
Should such an agent be developed, two types of experiments will be essential: 1) The agent should be trained on a large and diverse set of tasks to verify the absence of catastrophic forgetting. 2) The agent should be trained on a wide range of tasks with varying difficulties, followed by an evaluation of its capability to adapt to untrained tasks.

\paragraph{Applying Meta-Learning} In addressing the ARC tasks, the ability to apply learned concepts to entirely new contexts is essential. 
Meta-learning, or `learning to learn,' can be considered to adapt quickly to new tasks. 
Future studies should focus on developing meta-learning frameworks that can efficiently abstract underlying task structures, allowing for rapid generalization across varying problem settings. 
A deeper exploration into meta-learning could utilize frameworks such as Model-Agnostic Meta-Learning (MAML)~\cite{finn2017maml} which has shown promise in various domains.
Implementing and refining meta-learning techniques like MAML could lead to breakthroughs in developing AI models that can learn not only untrained tasks more efficiently but also their knowledge to solve untrained tasks.

\paragraph{Applying Transfer Learning} Transfer learning can significantly enhance the model's ability to utilize knowledge acquired in one context and apply it to different yet related tasks. This process often involves the use of deep neural networks, specifically their ability to approximate and adapt policies through layers that capture generalizable features, which are crucial for the successful transfer of knowledge across tasks within the same domain or between different domains ~\cite{taylor2009transfer}.
Conducting such studies would provide deeper insights into the flexible application of acquired knowledge, a critical aspect of analogical reasoning. These investigations will pave the way for creating adaptable and efficient learning systems that thrive in dynamically changing environments.

\paragraph{Theoretical Analysis of Model-Based RL} Finally, our findings indicate that model-based RL can facilitate more efficient learning in analogical reasoning tasks compared to model-free approaches. 
The inherent capabilities of model-based methods to infer and generalize from limited data were particularly beneficial. 
However, our study also highlights the need for further investigation into the specific mechanisms through which these models store and retrieve task-specific patterns. 
A deeper understanding of these processes could inform the development of more robust model-based systems, enhancing their capability to handle a wider array of complex reasoning tasks.

%% file: 5_relatedworks.tex
\section{Related Works}

The winner of the ARC Challenge~\cite{kaggle2020arcchallenge} utilized a total of 142 domain-specific languages (DSLs) to combine various transformed images~\cite{icecuber2020dsl}.
The ARC Challenge later expanded into ARCathon~\cite{lab422024arcathon}, and the ARCathon 2022 winner's approach involved exploring using 166 DSLs that included more complex and diverse features~\cite{hodel2023dsl}.
This attempted to challenge the broad generalization of DSLs; however, it also introduced limitations due to the inclusion of very infrequently used features, making the DSLs too complex.
Compared to humans can solve about 80\% of ARC evaluation tasks~\cite{johnson2021fast}, DSL-based search algorithms have shown performances around 30-40\%~\cite{bober2024ARClevel, lab422024arcathon}.
However, these approaches, being fundamentally simple search-based, have inherent weaknesses in complex tasks and are prone to overfitting to specific tasks defined by artificially designed DSLs, making them difficult to apply or generalize to other tasks. 
This fundamental limitation may render them unsuitable for analogical reasoning.

Subsequent studies have attempted to solve tasks using neural networks. 
Some research efforts were based on program synthesis, progressively recognizing increasingly complex patterns, while other studies have tried to learn the complexities of ARC through the computational abilities of Large Language Models (LLMs).
However, studies without DSLs have shown relatively low performance. 
Even the most recent studies based on program synthesis~\cite{ainooson2023approach} or LLMs~\cite{mirchandani2023large} have only achieved performances of 6.5--6.75\% on untrained tasks~\cite{butt2024codeit}.
Recently, there was interesting research utilizing human-solving processes~\cite{park2023unraveling}, but this study did not contain the performance on untrained tasks.
The low performance of these studies is attributed to the following reasons: In the case of program synthesis, the method of combining all possessed knowledge to acquire new knowledge leads to poor learning efficiency~\cite{ellis2021dreamcoder}, and LLMs are known to be weak in incremental reasoning~\cite{lee2024reasoning}. 
These limitations can be particularly harmful in evaluation tasks, which are known to be comparatively more complex and difficult than training tasks~\cite{bober2024ARClevel}. 
Additionally, the research utilizing human-solving processes has a risk to untrained tasks due to their reliance on offline learning from given solutions~\cite{tan2023large}.

Ensemble-based research that combines the previous approaches has also been announced~\cite{bober2024ARClevel, butt2024codeit}. 
These methods were expected to compensate for the weaknesses of existing approaches through the ensemble. 
An ensemble-based study combining program synthesis and LLMs showed a performance of 14.75\%~\cite{butt2024codeit}, and the approach of the ARCathon 2023 winner exhibited about 33\% performance, significantly surpassing previous studies. 
However, the performance of ensemble-based research that combined DSL, program synthesis, and LLMs was only 40.25\%~\cite{bober2024ARClevel}, which is just 0.25\% higher than the performance of a baseline that used only DSL. 
This result means that just one more task out of 400 evaluation tasks was solved. 
Ultimately, the experimental results of these ensemble-based studies further highlighted the limitation that the presence of program synthesis and LLMs does not significantly impact performance compared to DSL.

Meanwhile, there are some attempts to apply reinforcement learning (RL) algorithms in a limited way based on program synthesis and LLMs methods~\cite{butt2024codeit, alford2021neural}. 
The limited application of RL algorithms can be attributed to factors such as the absence of appropriate rewards, high-dimensional complex states, and extensive search spaces, but fundamentally, the absence of an environment for ARC was the main reason. 
However, an environment for training ARC tasks~\cite{lee2024arcle} has been developed recently. 
Furthermore, this research demonstrated successful training of an ARC task using PPO, suggesting the potential applicability of various RL models.

%% file: 6_conclusion.tex
\section{Conclusion}

In this paper, we compared analogical reasoning abilities between model-based and model-free RL, using a subset of ARC to demonstrate the effectiveness of model-based RL. 
This approach has shown superior learning efficiency and adaptability in tasks compared to model-free methods like PPO. 
Specifically, DreamerV3 outperformed in tasks similar to those it had previously solved, displaying a remarkable ability to generalize across various task dimensions.

The observed performance dips followed by rapid recoveries during DreamerV3’s training are thought to represent periods of conceptual consolidation. 
These intervals align with theoretical expectations of analogical reasoning in AI systems and highlight DreamerV3's potential to manage complex cognitive tasks. 
Future research should further explore these learning dynamics to better understand the mechanisms that enable such advanced reasoning. 
This will help enhance RL agents capable of analogical reasoning and adaptation in dynamic environments, reducing the gap between the reasoning ability of humans and RL agents.